\documentclass{article}
\usepackage{arxiv}
\usepackage[utf8]{inputenc} 
\usepackage[T1]{fontenc}    
\usepackage{hyperref}       
\usepackage{url}            
\usepackage{booktabs}       
\usepackage{amsfonts}       
\usepackage{nicefrac}       
\usepackage{microtype}      
\usepackage{lipsum}
\usepackage{float}
\usepackage{graphicx}
\usepackage{multirow}
\usepackage{array}
\usepackage{comment}
\usepackage{longtable}
\usepackage{ragged2e}
\usepackage{subfig}

\title{The development of a portable elbow exoskeleton with a Twisted Strings Actuator to assist patients with upper limb inhabitation}

\author{
  Rupal Roy  \\
  Mechatronics Engineering\\
  International Islamic University Malaysia\\
  Kula Lumpur, Malaysia\\
  \texttt{rupal.roy@live.iium.edu.my} \\
   \And
     MM Rashid   \\
  Mechatronics Engineering\\
  International Islamic University Malaysia\\
  Kula Lumpur, Malaysia\\
  \texttt{mahbub96@gmail.com} \\
   \And
  Md Manjurul Ahsan \\
  Industrial and Systems Engineering\\
  University of Oklahoma\\
  Norman, Oklahoma-73071 \\
  \texttt{ahsan@ou.edu} \\
   \And
 Zahed Siddique \\
  Department of Aerospace and Mechanical Engineering\\
  University of Oklahoma\\
  Norman, Oklahoma-73071\\
  \texttt{zsiddique@ou.edu}} 

\begin{document}
\maketitle

\begin{abstract}
Over the years, the number of exoskeleton devices utilized for upper-limb rehabilitation has increased dramatically, each with its own set of pros and cons. Most exoskeletons are not portable, limiting their utility to daily use for house patients. Additionally, the huge size of some grounded exoskeletons consumes space while maintaining a sophisticated structure and require more expensive materials. In other words, to maintain affordability, the device's structure must be simple. Thus, in this work, a portable elbow exoskeleton is developed using SolidWorks to incorporate a Twisted Strings Actuator (TSA) to aid in upper-limb rehabilitation and to provide an alternative for those with compromised limbs to recuperate. Experiments are conducted to identify the optimal value for building a more flexible elbow exoskeleton prototype by analyzing stress, strain conditions, torque, forces, and strings. Preliminary computational findings reveal that for the proposed intended prototype, a string length of.033 m and a torque value ranging from 1.5 Nm to 3 Nm are optimal.
\end{abstract}

\section*{Abbreviations}
TSA \quad Twisted Strings Actuator\\
ADL \quad Activities of Daily Living\\ 
ABS 	\quad Acrylonitrile Butadiene Styrene\\
DC	\quad	Direct Current\\
 PPR 	\quad	Pulse per Revolution\\

\section{Introduction}
Neuromuscular disease, such as stroke, is frequently the causes limb paralysis, which could have a substantial impact on the patients' activities of daily living (ADL), as the limbs could become completely paralyzed without appropriate training~\cite{fukuda2014effectiveness,kristensen2021neuromuscular}. However, this can be avoided if patients are treated from the onset. Conventional therapy, on the other hand, are quite time consuming. This is because repetitious tasks can be physically and mentally tiring, even more so for the physiologists or caretakers in charge~\cite{nguyen2018virtual}. In order to address such issues, robotics technology in rehabilitation has advanced to aid patients with motor impairments.\\
The number of robotic devices created to aid in upper limb rehabilitation has grown exponentially over the years~\cite{roy2021investigation}. Apart from regularly performing repeated duties on patients with high precision and accuracy, the robots were tailored to meet certain situations~\cite{hu2011advanced}. Thus, robot-assisted training has made a significant contribution to stroke recovery by giving patients with the necessary training. Most exoskeletons that have been developed are large, heavy, and grounded, making them difficult to pack and utilize~\cite{young2016state}.\\
Complete training for upper limb rehabilitation includes both active and passive types of rehabilitation, depending on the patient's condition. For patients with acute stroke, an active rehabilitation mode is used in which the robot directs patients who are unable to control their limbs to move their arms in the required motion pattern~\cite{vitiello2012neuroexos,zhang2018system}. As a result, patients who have suffered an acute stroke will undertake to active rehabilitation. Meanwhile, patients in passive rehabilitation can control some of their limbs on their own but will require assistance to move. 
There are two types of passive rehabilitation in which a supporting force are applied to the elbow during mobility in moderate stroke patients. It acts as a resistive force on the elbow during the entire recovery period~\cite{manna2017mechanism,teasell2020canadian}.\\
The elbow exoskeletons are being created to aid in the rehabilitation process for people who have lately suffered from elbow joint impairments. Thus, fundamental knowledge about the elbow must be learned in order to facilitate rehabilitation through the creation of an elbow exoskeleton. The elbow joint is composed of many bones such as humerus, radius, and ulna, that are linked at one end. While the humerus is positioned in the upper arm, the radius and ulna are in the forearm, which enables the arm to flex and extend around the elbow joint, as well as the forearm and wrist to pronate and supinate. Therefore, the range of motion of an elbow in flexion extension in a healthy person is 140 degrees~\cite{meislin2016comparison}. In comparison, the forearm pronation-supination angle is 160 degrees~\cite{schuind1991distal}, with 80 degrees for pronation and another 80 degrees for supination~\cite{malagelada2014elbow}.
The actuator's motion can be sent to a specific part of a structure via a transmission system. This mechanism affects motion by slowing it down and increasing the specified torque or force values, which is acceptable given that speed is not a concern for rehabilitation robots~\cite{pehlivan2012design,xiloyannis2019physiological}.\\
Recently, Twisted Strings Actuator (TSA) is introduced as a simple and small actuator~\cite{jeong2021applications}. As a result, the TSA is included into the proposed design due to its simplicity and light weight, in addition to creating an extraordinarily high transmission ratio for the joints to move. Additionally, the actuator attached to the strings can impart rotational motion on the strings, resulting in their length being reduced and providing a pulling force that results in linear motion for the load~\cite{park2014impedance}. Therefore, TSA may be considered with a nonlinear gear ratio due to the significant linear force generated by a low torque twisted string~\cite{popov2013bidirectional,seong2020development}.\\
In this artificial intelligence era~\cite{ahsan2022industry}, it is necessary to develop a more advanced and flexible elbow exoskeleton for user satisfaction and further improvement based on the user’s feedback. Additionally, the recent COVID-19 crisis makes it difficult to take the treatment from the hospital as well~\cite{ahsan2021detection,ahsan2021detecting,ahsan2020covid,ahsan2020deep}. Considering this opportunity into account, in this work, we have developed a portable elbow exoskeleton with twisted strings actuators that might assist the patients with upper limb inhabitation. 
The rest of the section is constructed as follows: in Section~\ref{method} materials and methods were described. Section~\ref{result} details our study findings and discusses overall results. Finally, Section~\ref{conclusions} summarizes the overall findings.
\section{Materials and methods}\label{method}
\subsection{Mechanical Design}
\begin{figure}[htbp]
    \centering
    \includegraphics{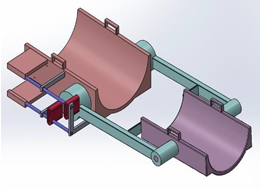}
    \caption{Design of the prototype using SolidWorks}
    \label{fig:fig1}
\end{figure}

SolidWorks 2017 software is used to design the elbow exoskeleton, which supports and moves the elbow joint in flexion-extension mode. The prototype as shown in Figure~\ref{fig:fig1}, is intended for patients who have suffered an acute stroke, with an emphasis on active rehabilitation, in which the actuator regulates elbow movements without involving muscular movements. A yoke-pin system has been used, in which the strings pull the yoke, rotating the link as they do so. By connecting two TSA modules in parallel, the contracting and extending of the exoskeleton will resemble the bicep and triceps, resulting a more comfortable training experience for the patients.
\\
The motor was selected through calculations of the torque needed to lift the forearm. Considering the maximum mass of a person’s forearm to be assumed as 2.5 kg and the length from the elbow joint to the center of the forearm to be 0.1 m~\cite{tsuji1995human}, the torque involved to move the arm can be calculated using the general formula of torque of a pivot due to gravity using Equation 1:
\begin{equation}
    \tau = mgd
\end{equation}
Where m is the mass of the object, g is the gravity constant, and d is the distance on the link from m is the center of rotation. The needed torque is determined to be 2.77 Nm. As a result, the GM37-520 12V DC Magnetic Encoder Gear Motor with a basic signal of 11 PPR (Pulse per Revolution) and a motor torque of 3 Nm is chosen.
\subsection{Controller Design}
For robotic systems, a feedback loop is considered as the primary control. As the controller controls the DC motor, the elbow exoskeleton may be rotated using the TSA. Simultaneously, the encoder considers the location of the motor shaft. As a result, anytime an interrupt is applied, the device will discontinue the operation after one cycle, acting as an emergency button as illustrated in Figure~\ref{fig:fig2} below.
\begin{figure}
    \centering
    \includegraphics{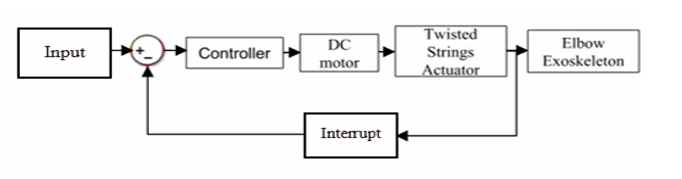}
    \caption{Controller Design}
    \label{fig:fig2}
\end{figure}
\subsection{Electrical Circuit Schematic Design}
The DC motor is connected to the strings via a connector that designed specifically in this experiment and is connected to the Arduino UNO via the L298P 2A motor drive shield. The motor drive shield receives 12V from a 12V power supply, which is sufficient to operate the motor. Additionally, the Arduino is connected to a Bluetooth module. After synchronizing the HM-10 Bluetooth module with the phone, the user can control the exoskeleton. As illustrated in Figure~\ref{fig:fig3}, the schematic design is as follows:
\begin{figure}[htbp]
    \centering
    \includegraphics{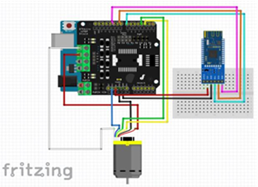}
    \caption{Schematic of the circuit}
    \label{fig:fig3}
\end{figure}
\subsection{Material and Strings Selections}
The prototype is developed using a 3D printer to save time and energy while yet delivering a lightweight material for the exoskeleton. The developed SolidWorks designed was saved in STL format, and from there, a layer-by-layer version of the design was created. The majority of the parts are made of a white ABS (Acrylonitrile Butadiene Styrene) plastic with a 200-micron thickness and 15\% infill. According to Mehmood et al. (2015), braided threads can support a greater load, which is perfect for this design~\cite{mehmood2015rotational}. Additionally, because braided strings are more robust, they are easier to use and maintain than non-braided strings. As a result, the braided type of string was chosen for this research. Due to the budget constraints, in this work we have employed braided cotton twine string as it is less expensive and stronger than the fishing string. While the cotton twine string may appear a little unkempt after a few uses, it is constructed with durable cotton that is resistant to tearing and has a long service life. The string has a diameter of 2mm. Due to the fact that the yoke has two sides, one antagonist and one protagonist, it has been determined that both the top and bottom strings should be 3.5 cm in length from the motor connector to the yoke.
\section{Results and discussion}\label{result}
\subsection{SolidWorks Simulation}
\begin{figure}
    \centering
    \includegraphics{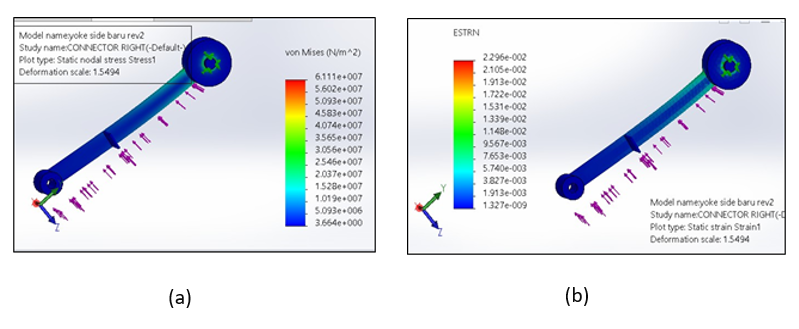}
    \caption{(a) Stress and (b) Strain result simulation}
    \label{fig:stress}
\end{figure}

On one of the connecting links, a SolidWorks simulation was performed. The test was conducted to observe how the link would react when subjected to a 3 Nm torque. Figure~\ref{fig:stress} (a) illustrates the stress result, which is related to the material type and weight. As indicated previously in the earlier section, the 3D print material is ABS, which appears to have low stress values. Stress is generated when a force is applied to a material, causing it to deform. Low stress readings indicate that the design is sufficiently robust. The strain consequence of bending is depicted in the accompanying Figure~\ref{fig:stress} (b). Strain can be viewed as the body's response to applied stress, and Figure~\ref{fig:stress} (b) demonstrates that the connection is far from possible damage. 
\subsection{Torque Analysis}
From the specifications of the DC motor as listed before, the torque provided by the motor can be calculated using Equation 2 as follows:
\begin{equation}
    P = \tau_{m}\omega
\end{equation}
where $P$ is the power, $\tau$ is the torque and $\omega$ is the angular velocity. Therefore, the motor torque, $\tau_m$ is found to be 2.92 Nm. The graph in Figure 5 shows each of the motor torque for the specific speed of the motor available from the same brand.
\begin{figure}[htbp]
    \centering
    \includegraphics{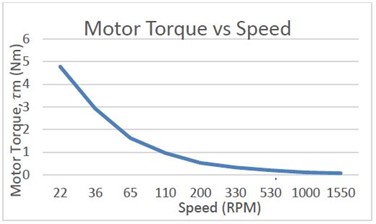}
    \caption{Torque required to lift forearm vs mass of forearm graph}
    \label{fig:my_label}
\end{figure}
By referring to the motor torque calculated here, and the torque required, it can be concluded that the torque provided by the motor chosen is the best and appropriate enough to lift the forearm using the exoskeleton.
\subsection{Force Analysis}
\begin{figure}[htbp]
    \centering
    \includegraphics{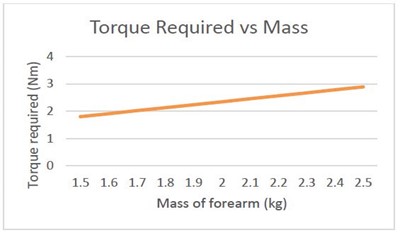}
    \caption{Torque required to lift forearm vs mass of forearm graph}
    \label{fig:torq}
\end{figure}

Since the torque provided by the motor is calculated to be 2.92 Nm, the twisting angle can be derived from the formula for torque required by the motor to provide a pulling force by means of TSA~\cite{mehmood2015rotational} as shown in Equation 3.
\begin{equation}
    \tau_{m} = \frac{F_{m}\theta r^2}{\sqrt{L^2-\theta^2r^2}}
\end{equation}
where $F_{m}$ is the force which TSA applies on the yokes and consequently the pins, $\theta$ is the twisting angle, L is the initial length of the strings and r is the radius of the strings.
The force which TSA applies on the yokes and consequently the pins, $F_{m}$  can be calculated by using Equation 4:
\begin{equation}
    F_{m} =\frac{F_t}{sin\beta}
\end{equation}
Meanwhile, the tangential force, $F_t$ acting on the pin caused by external load can be calculated by using Equation 5:
\begin{equation}
    F_{t} = \frac{mgl_ecos\beta}{r_pln}
\end{equation}
From the calculations, it can be found that $\theta$ is equal to 69.7 degree. Using different masses of the forearm, Figure~\ref{fig:torq} illustrated the graph of the required torque to lift the forearm against the mass using the above equation whereby it can be deduced that the torque required to lift the forearm is within 1.5 Nm to 3 Nm range.
\subsection{String Analysis}
\begin{figure}
    \centering
    \includegraphics{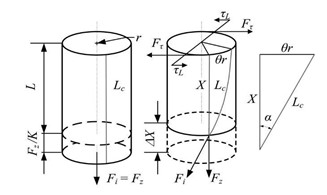}
    \caption{Twisted strings geometry~\cite{gaponov2013twisted}}
    \label{fig:fig7}
\end{figure}

\begin{figure}[htbp]
    \centering
    \includegraphics{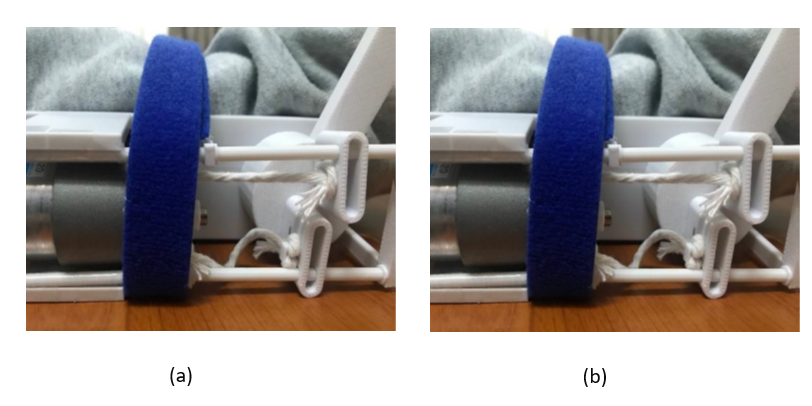}
    \caption{Condition of the strings during (a) extension and (b) flexion}
    \label{fig:fig8}
\end{figure}

\begin{figure}
    \centering
    \includegraphics{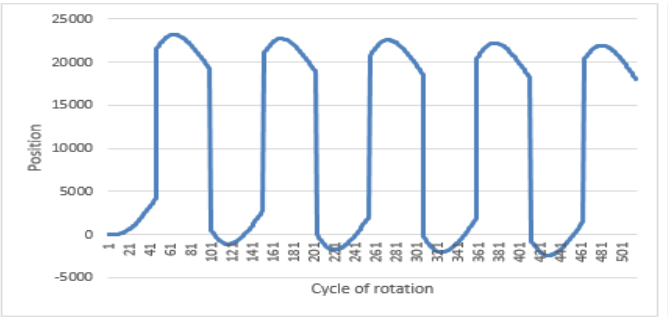}
    \caption{Position of motor per cycle}
    \label{fig:fig9}
\end{figure}

\begin{figure}
    \centering
    \includegraphics{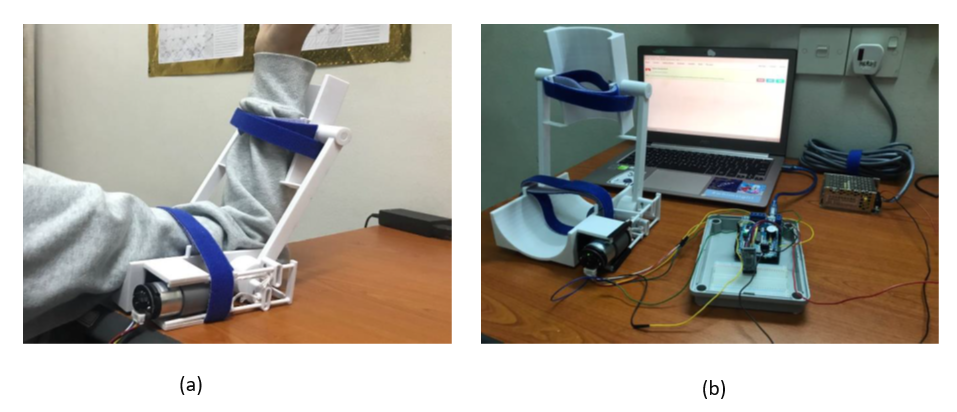}
    \caption{Assembled prototype of the (a) elbow exoskeleton and (b) prototype setup}
    \label{fig:fig10}
\end{figure}

The condition of the strings is analyzed and the geometry of the twisting effect on the string can be observed in Figure~\ref{fig:fig7}.
While twisting, the contracted length of the loaded string can be calculated from the formula using Equation 6:
\begin{equation}
   X = \sqrt{L^2_e-\theta^2r^2_o} = L_ecos\alpha 
\end{equation}
where $L_e$ is the extended length of the loaded string, $r_o$ is the radius of the string, $\theta$ is the angle of twisting and $\alpha$ is the helix angle. Since the extended length of the loaded string for both the top and bottom parts of the yoke is 0.035 m, the contracted length of that string is calculated as 0.033 m.\\
The operation of the system requires an installation of the mobile application and connect it to the Bluetooth module. Once connected, the user may activate the device by pressing ACTIVATE button. For safety purposes, the DEACTIVATE button is also added and once the user press it, the device will complete one cycle before it stops. If the user does not deactivate manually, the exoskeleton will stop after 5 cycles which will help the patient to prevent from hurting their arm from overworking. For each cycle, the motor will first rotate clockwise for 3 seconds and then counterclockwise for another 3 seconds. Then, the motor will stop for 5 seconds before continuing the next cycle. Once the 5 cycles are finished, the user needs to activate it again.\\
According to Saeed et al. (2013), it was acknowledged that the ROM (Range of motion) needed for an elbow to move in performing ADL is from 30 degrees to 130 degrees, hence the exoskeleton is designed to rotate within a 50 degrees’ angle~\cite{saeed2013modelling}. This is because the 0 degree starts from the forearm’s supporting pad and the links to be 90 degrees from upper arm’s supporting pad. As the top string twists and contracts the length while the bottom string extends, the link would go up to 50 degrees of angle and vice versa. The positions of the strings can be seen as in Figure~\ref{fig:fig8}.
The encoder will read the positions from the clockwise and counterclockwise rotations of the motor and know whether the device is working accordingly or not. The positions of the motor for 5 full cycles as recorded by the encoder is illustrated in Figure~\ref{fig:fig9}.
The fabricated and assembled prototype of the design can be seen in Figure~\ref{fig:fig10} (a). The prototype setup can be seen in Figure~\ref{fig:fig10} (b) where the circuit box is attached to the elbow exoskeleton by adopting the style of a blood pressure machine.
\section{Conclusions}\label{conclusions}
With the help of a Twisted Strings Actuator (TSA), an elbow exoskeleton is developed in this study. Exoskeleton elbows may now be activated using the TSA, which has undergone extensive research and development. Finally, all the goals have been met, including designing a portable exoskeleton that is lightweight and comfortable to wear, understanding the concept and use of TSA, and integrating it into the exoskeleton to create a simple and economical actuator. It was found that using software and practical evaluation, it was possible to test the practicality of the design. As a lightweight ungrounded exoskeleton, the elbow exoskeleton was constructed using carefully selected materials. Future works includes improving the proposed prototype based on the user experience, developed more flexible and cheaper elbow exoskeleton considering different materials, analyze the patients data using machine learning approaches~\cite{ahsan2021effect}, and integrated the prototype with artificial intelligence systems.
\bibliographystyle{unsrt}  
\bibliography{main}

\end{document}